\documentclass[10pt,twocolumn,letterpaper]{article}

\usepackage{cvpr}              % To produce the CAMERA-READY version
\usepackage{graphicx}
\usepackage{amsmath}
\usepackage{amssymb}

\usepackage{booktabs}
\usepackage[accsupp]{axessibility}
\graphicspath{{images/}}

\usepackage[pagebackref,breaklinks,colorlinks]{hyperref}

\usepackage[capitalize]{cleveref}
\crefname{section}{Sec.}{Secs.}
\Crefname{section}{Section}{Sections}
\Crefname{table}{Table}{Tables}
\crefname{table}{Tab.}{Tabs.}

\def\cvprPaperID{1} % *** Enter the CVPR Paper ID here
\def\confName{CVPR}
\def\confYear{2022}

\begin{document}

%%%%%%%%% TITLE - PLEASE UPDATE
%\title{\LaTeX\ Author Guidelines for \confName~Proceedings}
\title{Context Attention Network for Skeleton Extraction}

% \author{Zixuan Huang, Yunfeng Wang, Zhiwen Chen, Xin Gao, Ruili Feng, Xiaobo Li \\
% Alibaba Group, University of Science and Technology of China \\
% \{\tt\small zixuan.huangzixuan, weishan.wyf, zhiwen.czw, zimu.gx\}@alibaba-inc.com, ruilifengustc@gmail.com, \\
% \tt\small xiaobo.lixb@alibaba-inc.com
% }

% \author{Zixuan Huang\textsuperscript{1}, Yunfeng Wang\textsuperscript{1}, Zhiwen Chen\textsuperscript{1}, Xin Gao\textsuperscript{1}, Ruili Feng\textsuperscript{2}, Xiaobo Li\textsuperscript{1} \\
% \textsuperscript{1}Alibaba Group, \textsuperscript{2}University of Science and Technology of China \\
% \{\tt\small zixuan.huangzixuan, weishan.wyf, zhiwen.czw, zimu.gx\}@alibaba-inc.com, ruilifengustc@gmail.com, \\
% \tt\small xiaobo.lixb@alibaba-inc.com
% }

\author{Zixuan Huang\textsuperscript{1}, Yunfeng Wang\textsuperscript{1}, Zhiwen Chen\textsuperscript{1}, Xin Gao\textsuperscript{1}, Ruili Feng\textsuperscript{2}, Xiaobo Li\textsuperscript{1} \\
[0.2cm]\textsuperscript{1}Alibaba Group \quad
\textsuperscript{2}University of Science and Technology of China \\
\{\tt\small zixuan.huangzixuan, weishan.wyf, zhiwen.czw, zimu.gx\}@alibaba-inc.com, ruilifengustc@gmail.com, \\
\tt\small xiaobo.lixb@alibaba-inc.com
}

% \author{
% Yuqing Wang$^1$, 
% %
% Zhaoliang Xu$^1$, 
% %
% Xinlong Wang$^2$,
% %
% Chunhua Shen$^2$, 
% %
% Baoshan Cheng$^1$,
% %
% Hao Shen$^1$\thanks{Corresponding author.}, 
% %
% %
% Huaxia Xia$^1$
% \\[0.2cm] 
% $ ^1$ Meituan 
%  ~
%  ~
%  ~
%  ~
%  ~
% %
% $ ^2$ The University of Adelaide, Australia
% %
% \\{\tt\small yuqingwang1029@gmail.com, shenhao04@meituan.com}
% }

% \author[1]{Zixuan Huang}
% \author[1]{Yunfeng Wang}
% \author[1]{Zhiwen Chen}
% \author[1]{Xin Gao}
% \author[2]{Ruili Feng}
% \author[1]{Xiaobo Li}

% \affil[1]{Alibaba Group}
% \affil[2]{University of Science and Technology of China}
% \affil[ ]{\{\tt\small zixuan.huangzixuan, weishan.wyf, zhiwen.czw, zimu.gx\}@alibaba-inc.com, ruilifengustc@gmail.com\\
% \tt\small xiaobo.lixb@alibaba-inc.com
% }

% , , , ,  \\
% ,  \\
% \{\tt\small zixuan.huangzixuan, weishan.wyf, zhiwen.czw, zimu.gx\}@alibaba-inc.com, ruilifengustc@gmail.com, \\
% \tt\small xiaobo.lixb@alibaba-inc.com
% }

\maketitle

%%%%%%%%% ABSTRACT
\begin{abstract}
Skeleton extraction is a task focused on providing a simple representation of an object by extracting the skeleton from the given binary or RGB image. In recent years many attractive works in skeleton extraction have been made. But as far as we know, there is little research on how to utilize the context information in the binary shape of objects.
In this paper, we propose an attention-based model called Context Attention Network (CANet), which integrates the context extraction module in a UNet architecture and can effectively improve the network’s ability to extract the skeleton pixels. Meanwhile, we also use some novel techniques including distance transform, weight focal loss to achieve good results on the given dataset. Finally, without model ensemble and with only 80\% of the training images, our method achieves 0.822 F1 score during the development phase and 0.8507 F1 score during the final phase of the Pixel SkelNetOn Competition, ranking 1st place on the leaderboard.

\end{abstract}

%%%%%%%%% BODY TEXT
\section{Introduction}
\label{sec:intro}
Skeleton extraction, also known as skeletonization, is a task focused on providing a simple representation of an object by extracting the skeleton pixels from the given binary or RGB image~\cite{liu2021adaptive}. Nowadays, skeleton extraction is widely used in many fields, including object recognition~\cite{yang2016object}, pose estimation~\cite{wei2016convolutional} and motion forecasting~\cite{wang2021simple}, etc. Traditional methods are usually divided into three categories: morphological thinning methods~\cite{zhang1984fast}, geometric methods, and distance transform-based methods~\cite{panichev2019u}. These classical methods usually provide low accuracy results and are sensitive to noise at the edge of the shape. In recent years, with the development of artificial intelligence, some deep learning-based skeleton extraction approaches have been proposed, which usually treat extracting skeleton pixels as a classification problem. These methods take the binary or RGB image as input and directly predict the skeleton pixels, which avoids complex post-processing. But how to improve the performance of deep learning-based methods is still a challenge. 

\begin{figure}

	\includegraphics[scale=.45]{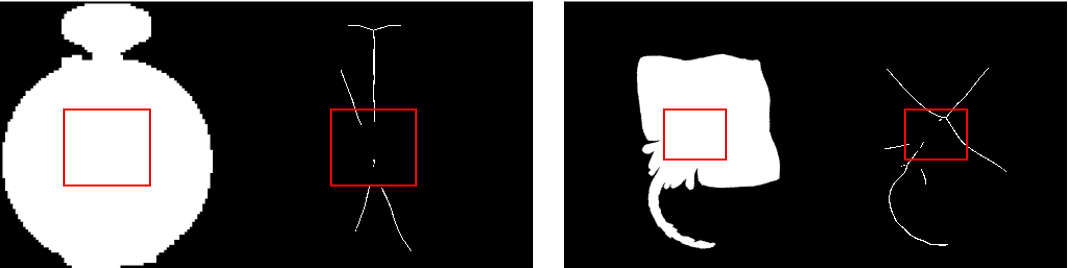}
    \caption{Badcase Analysis. For each subfigure, left is the binary shape input and right is the predicted skeleton of a basic UNet. Since there is no texture information in this scenario, structure information is crucial to final predictions. However, the center area (the red boxes) of binary shape image is lacking structure information, prone to broken and finely skeletons.}
    \label{fig:badcase}

\end{figure}

As shown in Figure~\ref{fig:badcase}, a typical skeleton extraction network fails to predict skeleton pixels in the plain area of the input shape image. On the one hand, the kernel size of convolutional layers is limited, thus the pixels in the plain area can't adopt the guidance from object shape. On the other hand, purely shape inputs lose most structure information in the object (except the boundaries), making it hard to determine the label of pixels in the center of objects.

In this paper, we propose Context Attention Network (CANet) to alleviate the aforementioned problems. In order to extract contextual cues efficiently, we modify the vanilla UNet with several vital updates, which is proved to be useful to improve the accuracy of skeleton extraction. Besides, we find that using the distance transform image that contains more information about object structure as input can ease the difficulty of skeleton extraction, further improving accuracy. With these novel optimizations, CANet outperforms existing methods and other participants on the Pixel SkelNetOn benchmark.

\section{Related Work}
Traditional methods of skeleton extraction are usually divided into three parts. For example, Zhang et al.~\cite{zhang1984fast} propose a fast parallel thinning algorithm, which iteratively removes the boundary and corner pixels of the object. After several iterations, only a skeleton of the object remains. Lu et al.~\cite{H1986A} improve this method by preserving necessary and essential structures which should not be deleted. However, these classical methods usually provide low accuracy results and are sensitive to noise at the edge.

Recently, many deep learning-based approaches have been proposed. Shen et al.~\cite{2017DeepSkeleton} propose a fully convolutional network, which is designed to extract the skeleton in different scales from multi stages. In~\cite{demir2019}, the authors use a vanilla pix2pix model with distance transformation preprocessing to extract the skeleton pixels. This preprocessing can reduce the learning difficulty of the network. Panichev et. al \cite{panichev2019u} introduce a U-Net based approach for direct skeleton extraction and get high performance. In~\cite{tang2021distance}, the authors continue to study preprocessing methods and propose to use Smooth Distance Estimation (SDE) and Edge Transformation to preprocess the input data, which wins the 1st place in the Pixel SkelNetOn 2021 Challenge. Nguyen~\cite{nguyen2021u} makes improvements to the original U-Net architecture using the attention mechanism and exploiting the auxiliary tasks, which also got excellent performances.

\section{Method}
In this section, we will introduce our main method and the tricks used to improve model performance, which contains network design, data processing, and loss strategies.
%强调：只使用了80%的数据，单模型达到0.822
\subsection{Network Design}
\begin{figure*}
\centering
	\includegraphics[scale=.4]{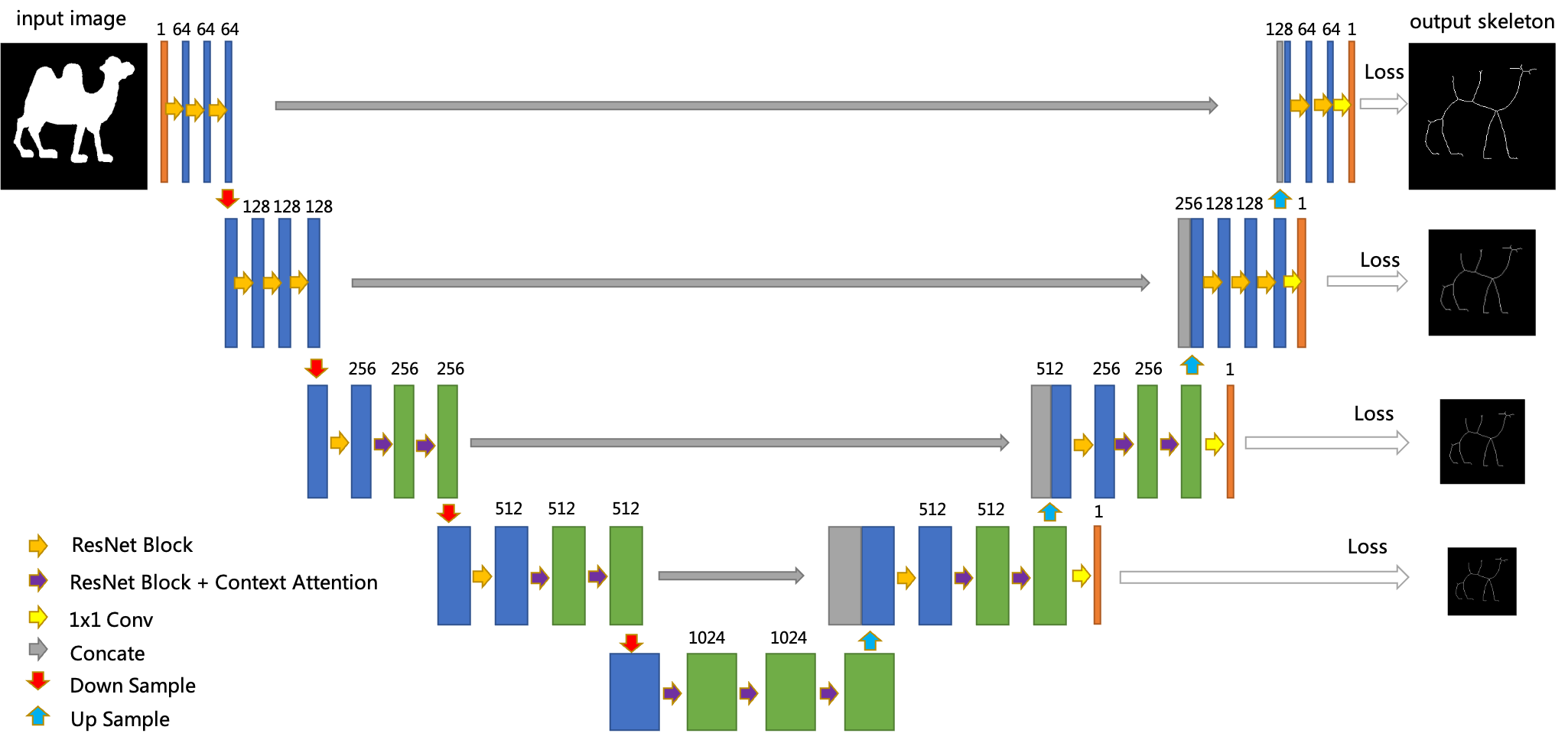}
    \caption{Context Attention Network: Our network is based on an encoder-decoder structure. Each stage of the encoder and decoder has a sequence of 3 ResBlocks. We add context attention block at stages 3-5 of the network. At the end of each encoder stage, max pooling is applied. The decoder stage contains a transposed convolutional layer, concatenated with a corresponding output of the encoder stage. At the end of the encoder stage, a $1 \times 1$ convolution is applied to generate skeleton predictions.}
    \label{fig:fig1}

\end{figure*}

\begin{figure}
\centering
	\includegraphics[scale=.3]{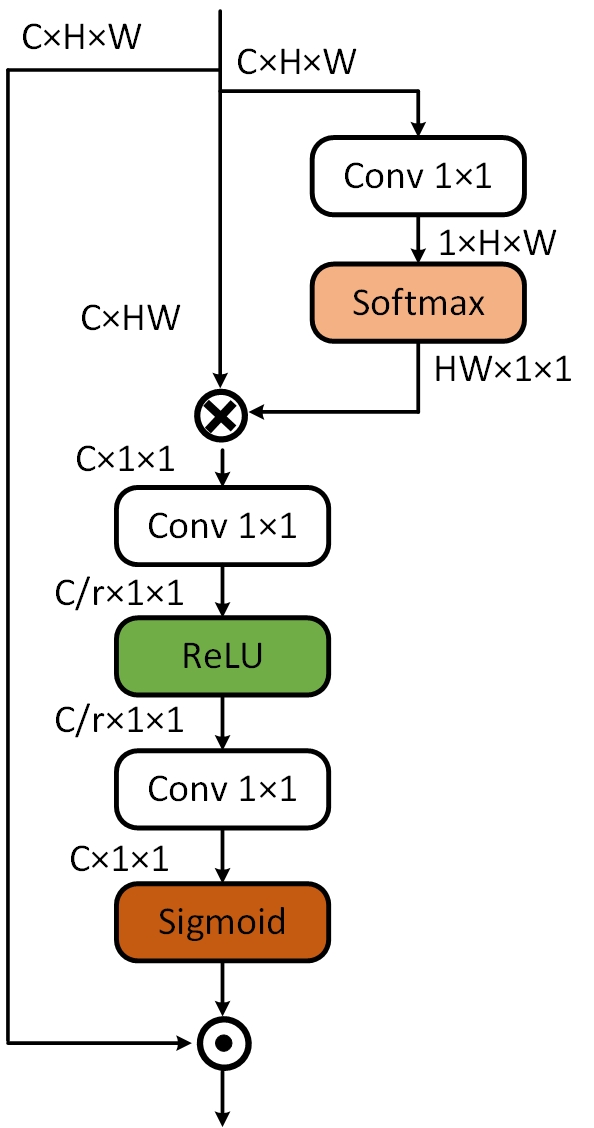}
    \caption{Context Attention Block: This Block is combined with spatial attention and channel attention. Firstly, matrix multiplication is adopted to acquire the correlation between every position of the feature map. Then we use channel attention to reassign channel weight.}
    \label{fig:fig2}

\end{figure}

Following~\cite{demir2019}, we treat the skeleton extraction as a pixel-wise binary classification problem. The input is binary images without texture, which contains less information and can be regarded as a low-level task. Therefore, we use UNet network~\cite{ronneberger2015u} as the baseline model. Based on this vanilla architecture, we make several optimizations to improve its learning capabilities. The model structure is shown in Figure~\ref{fig:fig1}.

\noindent \textbf{Residual Block.} In order to improve the effectiveness of the model, we replace the dual convolutional block (DualConvBlock) in UNet with the residual block (ResBlock)~\cite{he2016deep} since the residual structure can integrate semantic features of different convolutional layers while preventing the model from over-fitting. Specifically, we replace each DualConvBlock with 3 ResBlocks in all stages of encoder and decoder. 

\noindent \textbf{Context Attention Block.} In the neural network design, the attention module has become a powerful architecture to improve the model effect. SENet~\cite{hu2018squeeze} adaptively recalibrates channel-wise feature responses by explicitly modeling interdependencies between channels. Non-local~\cite{wang2018non} adopt spatial attention module to acquire the correlation between every position of the feature map. Inspired by these ideas, the Context Attention block contains spatial and channel attention simultaneously. As shown in Figure~\ref{fig:fig2}, the long-range space dependence is captured by matrix multiplication~\cite{wang2018non}. Then we use channel attention to reassign channel weight. ReLU function is added to increase the non-linear of the model~\cite{hu2018squeeze}. At the same time, in order to reduce the calculation of the model, the number of channels in the middle convolutional layer is reduced to 1/16. Applying Context Attention Block at the beginning of the network hurts the performance, thus we only deploy it in the higher stages(3-5).

\subsection{Data Processing}
% 输入，mask-> dt, 空洞填补，dt+ edge分数低，然后去掉edge，只使用dt 

\begin{figure}
\centering
	\includegraphics[scale=.45]{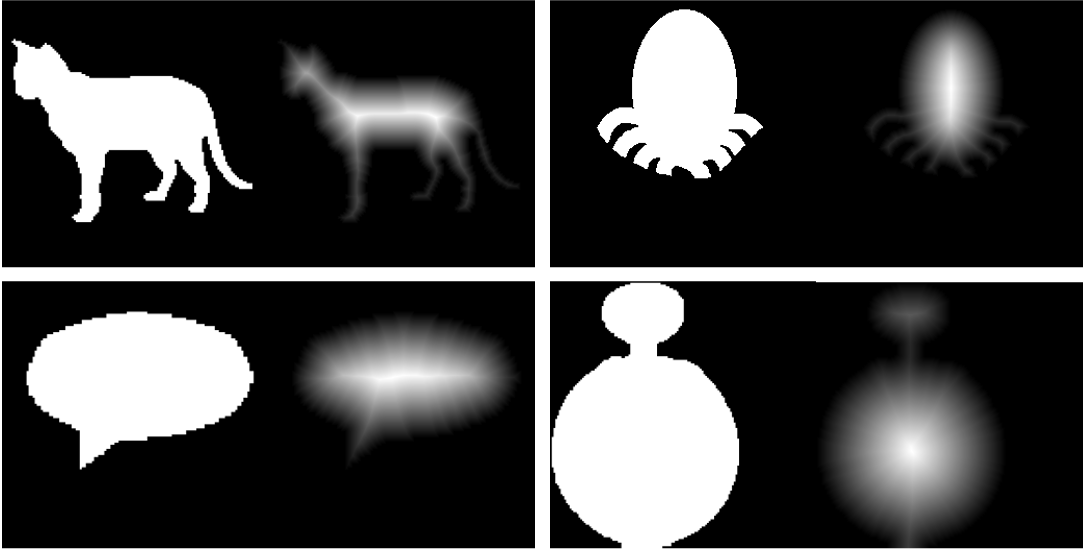}
    \caption{Comparison of the binary shape image and the output of the distance transform. Pixels in the latter are more distinguishable, making the skeleton extraction easier.}
    \label{fig:shape_vs_dt}
\end{figure}

Initially, the binary shape image of the object is used as the input. However, as shown in Figure~\ref{fig:shape_vs_dt}, we find that the output of the distance transform applied to the shape image contains more intuitive information closer to the skeleton of the object and can reduce the learning difficulty of the network. Meanwhile, we find that there are some holes in the output of the distance transform. The prediction has been significantly improved after filling the holes in the image. Then we try to concatenate the original image, the image with distance transform and the outline extracted by canny algorithm~\cite{1986A}, but unfortunately, the score is lower than expected. Following~\cite{tang2021distance}, We also try to replace distance transform with soft distance transform (SDE) but do not have a good result.

\subsection{Loss Design} 

%，辅助loss
Following~\cite{nguyen2021u}, Auxiliary loss is adopted to boost the performance of the model. Feature maps with different resolutions extracted from each stage of the decoder are fed into a $1 \times 1$ convolutional layer to get low-resolution prediction results. The losses calculated from low-resolution skeletons with down-sample ground truth are added to the final loss.

In terms of the loss function, we use the combination of Dice Loss~\cite{milletari2016v} and Focal Loss~\cite{lin2017focal} in our experiments to optimize the network because skeleton extraction is a data-imbalanced task.

Dice Loss is commonly used in semantic segmentation tasks, which is defined as below:

\begin{equation}
L_{dice} = 1 - 2\frac{\sum_{i}y_ip_i + \epsilon}{\sum_{i}y_i+\sum_{i}p_i+\epsilon},
\end{equation}
where $y_i$ is the target label, $p_i$ is the predict result and $\epsilon$ is a small constant to avoid division by zero (We set $\epsilon=1.0$ in all experiments).

Besides, we use weighted focal loss, which is defined as below:

\begin{equation}
L_{focal} = W_{pos} p^{\gamma}log(p)+ W_{neg}(1-p)^{\gamma}log(1-p),
\end{equation}
where $W_{pos}$ and $W_{neg}$ are the weights of positive and negative class respectively, $p$ is the probability that the sample belongs to positive class and $\gamma$ is the focusing parameter.

The final loss formula is:
\begin{equation}
L_{total} = \lambda_{dice} L_{dice} + \lambda_{focal} L_{focal},
\end{equation}

Where $\lambda_{dice}$ and $\lambda_{focal}$ are hyperparameters to balance the value of losses. We set $\lambda_{dice}=1.0$ and $\lambda_{focal}=100.0$.

\section{Experiments}
In this section, we first detail our experimental settings, followed by a comparison between our method and other methods. Then we conduct the ablation studies with different parameters. Finally, we visualize the results of our approach.
% 我们的实验结果，一步步的提升数据
% 和现有方案的指标对比
% 结果可视化
% 一些尝试过但不work的点（）
\subsection{Experimental Settings}

\noindent \textbf{Datasets.}
Our model is trained on the Pixel SkelNetOn Dataset provided by the SkelNetOn 2022 Challenge~\cite{demir2019}. Pixel SkelNetOn Dataset contains 1,725 binary images with resolution $256 \times 256$, which splits into 1,218 training images, 241 validation images and 266 test images. We divide the training dataset into the training set and the split-test set as a ratio of 80\%:20\%.

\noindent \textbf{Implementation Details.}
% 网络所用框架，学习率，训练周期，训练时间
The SGD optimizer with a learning rate of 0.02 and cosine annealing algorithm is used. We also use the F1 score to evaluate the model performance. Following~\cite{nguyen2021u}, adaptive threshold selection is adopted to select the best threshold on the split-test set.

\subsection{Main Results}
Our method is compared with participants shown on the Pixel SkelNeton leaderboard. As shown in Table~\ref{tab:vs_competitors} and in Table~\ref{tab:vs_competitors_final}, whether in development or final phase, our method outperforms current methods with a large margin. 
% Besides, our method also surpasses the champion solution of Pixel SkelNeton Challenge 2021.

\begin{table}[h]
\caption{Leaderboard of Pixel SkelNetOn Challenge (Development phase).}
\begin{center}
\begin{tabular}{c|c|c}
\hline
Rank & Team name & F1 score \\ 
\hline
\textbf{1} & \textbf{huangzixuan0508(Ours)} & \textbf{0.8220} \\
2 & neptuneai & 0.7972 \\
3 & lv.zf & 0.7935 \\
4 & \_likyoo & 0.7846 \\
5 & Young\_Ji & 0.7400 \\
6 & kyriemelon & 0.6745 \\
\hline
\hline
- & 1st Place of Pixel SkelNetOn, 2021 & 0.8129
\end{tabular}
\end{center}
\label{tab:vs_competitors}
\end{table}

\begin{table}[h]
\caption{Leaderboard of Pixel SkelNetOn Challenge (Final phase).}
\begin{center}
\begin{tabular}{c|c|c}
\hline
Rank & Team name & F1 score \\ 
\hline
\textbf{1} & \textbf{huangzixuan0508(Ours)} & \textbf{0.8507} \\
2 & Young\_Ji & 0.8359 \\
3 & lv.zf & 0.8333 \\
4 & jiliushi & 0.8299 \\
5 & \_likyoo & 0.8289 \\
\end{tabular}
\end{center}
\label{tab:vs_competitors_final}
\end{table}

\subsection{Ablation Study}
In this part, we record our attempts to improve model performance in different ways, including network design, input format, and loss design.
% and multi-model ensemble strategies. 
\subsubsection{Network design}
Table~\ref{table:network} shows the results of different network architectures. After adding ResBlock, the representation ability of the model is improved, thus the F1 score is raised from 0.788 to 0.801. When Context Attention Block is added to the deep stages of the model, high-level semantic features can be weighted in spatial and channel dimensions, which improves the skeleton extraction score. At the same time, we find that the results get worse when adding Context Attention Block to the shallow stages of the model. Usually, the shallow layers are used to extract low-level features without high-level semantic information, and this ability may be adversely affected by attention modules.

\begin{table}[h]
\caption{Ablation experiments on network architectures. ``RB'' means ``ResBlock'', ``CA'' means ``Context Attention''. ``1-5'' means stage1 to stage5, ``3-5'' means stage3 to stage5.}
\begin{center}
\begin{tabular}{l|c}
\hline
Network Architecture & F1 score \\ 
\hline
UNet & 0.788 \\
UNet + RB & 0.801 \\ 
% ResBlock + Dilation Conv & 0.800 \\ 
UNet + RB + CA 1-5 & 0.801 \\
% ResBlock + Context Attention_A & 0.801 \\
UNet + RB + CA 3-5 & \textbf{0.806} \\
\hline
\end{tabular}
\end{center}
\label{table:network}
\end{table}

\subsubsection{Data Processing}
%% Data
Table~\ref{table:input} shows the results of different input formats. Initially, we use 80\% training data of binary shape image and get a score of 0.806. We try to use the instance segmentation network~\cite{wang2020solov2} to predict the shapes of RGB images in Image SkelNetOn Challenge. 2630 images were selected and added to the training set. Unexpectedly, the score is decreased, due to the inconsistent data distribution from different datasets. In terms of the input data format, images processed by distance transform are used as the input with a score of 0.803. After the hole areas are repaired, the score is further improved to 0.822. Furthermore, we also try to concatenate binary images and the outputs of distance transform as input data, but the score is not improved. We also implement
the Soft Distance Transform used in~\cite{tang2021distance}, but the results get worse.
\begin{table}[h]
\caption{Ablation experiments on the input data format of the network.}
\begin{center}
\begin{tabular}{l|c}
\hline
Input data & F1 score \\ 
\hline
Shape & 0.806 \\
Shape + RGB & 0.777 \\ 
Distance & 0.803 \\
Shape + Distance & 0.794 \\
Repaired distance & \textbf{0.822} \\
Soft distance & 0.799 \\
\hline
\end{tabular}
\end{center}
\label{table:input}
\end{table}

% Loss
\subsubsection{Loss}
In this part, we show the influence of different loss functions. From Table~\ref{table:loss} we can find that combining Focal Loss and Dice Loss improves networks' accuracy. Besides, we find that the weights of positive and negative samples are important for skeleton extraction. Specifically, adjust $W_{pos}$ from default 50 to 0.01, $W_{neg}$ from 0.1 to 0.99 can further improve F1 score from 0.782 to 0.788.

\begin{table}[h]
\caption{Ablation experiments on loss strategies.}
\begin{center}
\begin{tabular}{l|c}
\hline
Loss & F1 score \\ 
\hline
Focal Loss ($W_{pos}$=50.0, $W_{neg}$=0.1) & 0.778 \\
Dice Loss & 0.770 \\
Dice Loss + Focal Loss ($W_{pos}$=50.0, $W_{neg}$=0.1) & 0.782 \\
Dice Loss + Focal Loss ($W_{pos}$=0.01, $W_{neg}$=0.99) & \textbf{0.788} \\
\hline
\end{tabular}
\end{center}
\label{table:loss}
\end{table}

\subsection{Visualization}
\begin{figure}[h]
\centering
	\includegraphics[scale=.45]{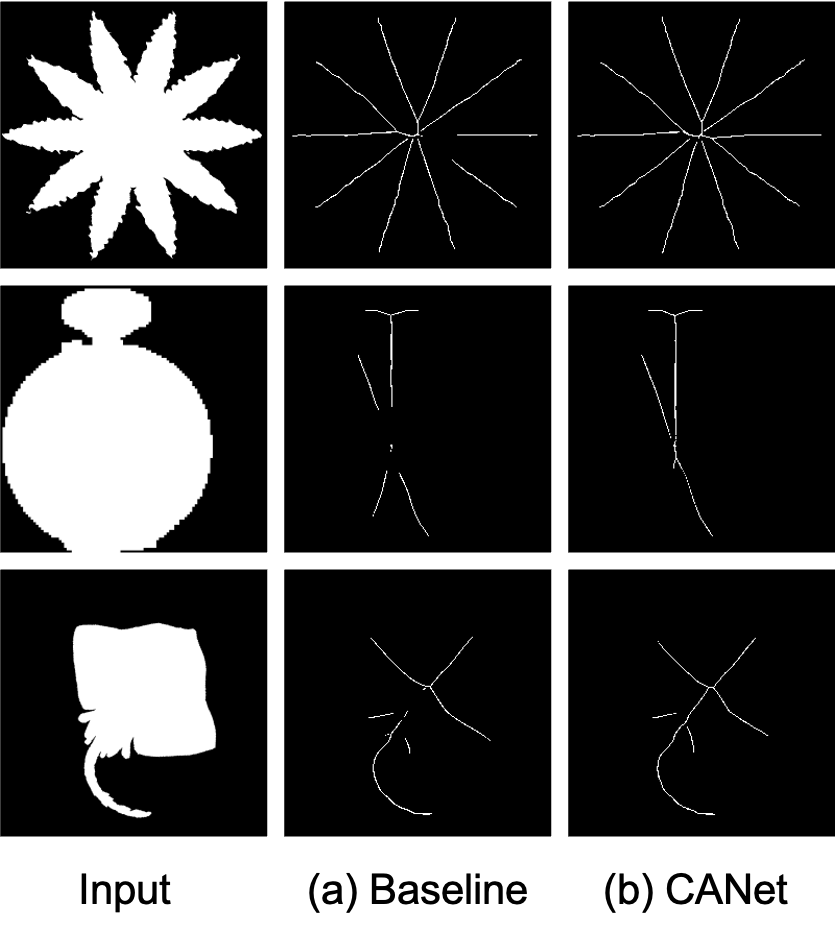}
    \caption{Qualitative comparison. Here, we compare CANet with the baseline on the validation dataset.}
    \label{fig:vs_baseline}
\end{figure}

%实验结果可视化对比
We visualize the predicted skeletons of our method and the baseline method in Figure~\ref{fig:vs_baseline}. With the help of techniques proposed in CANet, the skeletons of plain area can be recovered successfully.

\section{Conclusion}
In this paper, we propose the Context Attention Network for skeletonization. With only 80\% of the training data and a single model, our method achieves 0.822 and 0.8507 on the F1 score metric during the development and final phase of Pixel SkelNetOn Challenge, ranked as top-1 on the leaderboard. But in some cases, the problem of line breaks has not been well solved. In the next step, we will further explore the model structure and data processing method to reduce line breaks in the prediction process.

%%%%%%%%% REFERENCES
{\small
\bibliographystyle{ieee_fullname}
\bibliography{main.bib}
}

\end{document}